\documentclass{article} 
\usepackage{RobustML_iclr2021_conference,times}

\usepackage{amsmath,amsfonts,bm, bbm}

\def\eqref#1{equation~\ref{#1}}

\def\1{\bm{1}}

\def\mI{{\bm{I}}}

\DeclareMathAlphabet{\mathsfit}{\encodingdefault}{\sfdefault}{m}{sl}
\SetMathAlphabet{\mathsfit}{bold}{\encodingdefault}{\sfdefault}{bx}{n}

\newcommand{\R}{\mathbb{R}}
\newcommand{\one}{\mathbbm{1}}

\usepackage{hyperref}
\usepackage{url}
\usepackage{graphicx}
\usepackage{caption}
\usepackage{subcaption}
\usepackage{makecell}
\newcolumntype{C}[1]{>{\centering\arraybackslash}m{#1}}

\usepackage{color}

\title{Gi and Pal Scores: Deep Neural Network\\Generalization Statistics}

\author{Yair Schiff, Brian Quanz, Payel Das, Pin-Yu Chen \\
IBM Research, Yorktown Heights, NY 10598 \\
\texttt{\{yair.schiff,pin-yu.chen\}@ibm.com},\texttt{\{blquanz,daspa\}@us.ibm.com}\\
}

\iclrfinalcopy

\begin{document}

\maketitle

\begin{abstract}
The field of Deep Learning is rich with empirical evidence of human-like performance on a variety of regression, classification, and control tasks.
However, despite these successes, the field lacks strong theoretical error bounds and consistent measures of network generalization and learned invariances.
In this work, we introduce two new measures, the Gi-score and Pal-score, that capture a deep neural network's generalization capabilities.
Inspired by the Gini coefficient and Palma ratio, measures of income inequality, our statistics are robust measures of a network's invariance to perturbations that accurately predict generalization gaps, i.e., the difference between accuracy on training and test sets.
\end{abstract}

\section{Introduction}
Neural networks have produced state-of-the-art and human-like performance across a variety of tasks, from image classification to autonomous driving \citep{sengupta2020review}.
With this rapid progress has come wider-spread adoption and deployment.
Given their prevalence and increasing applications, it is important to better understand why neural nets are able to generate such high performance that often generalizes to unseen data and to estimate how well a trained net will generalize.

Various attempts at bounding and predicting neural network generalization are well summarized and analyzed in the recent survey \cite{jiang2019fantastic}.
While both theoretical and empirical progress has been made, there remains a gap in the literature for an efficient and intuitive measure that can predict generalization given a trained network and its corresponding data \textit{post hoc}.
Hoping to fill this gap, the recent Predicting Generalization in Deep Learning (PGDL) competition described in \cite{jiang2020neurips} encouraged participants to provide \textit{complexity} measures that would take into account network weights and training data to predict generalization gaps, i.e., the difference between performance on training and test sets.
In this work, we propose two new measures called the Gi-score and Pal-score that present progress towards this goal.
Our new statistics are calculated by measuring a network's performance on training data that has been perturbed with varying magnitudes and comparing this performance to an idealized network that is unaffected by all magnitudes of perturbation.

\section{Related work}\label{sec:related_work}
The PGDL competition resulted in several proposals of complexity measures that aim to bound and predict neural network generalization gaps.
While several submissions build off the work of \cite{jiang2018predicting} and rely on margin-based measures, we will focus on those submissions that measure perturbation response, specifically mixup, since this is most relevant to our work.
Mixup, first introduced in \cite{zhang2017mixup}, is a novel training paradigm in which training occurs not just on the given training data, but also on linearly interpolated points.
Manifold Mixup training extends this idea to interpolation of intermediate network representations \citep{verma2019manifold}.

Perhaps most closely related to our work is that of the winning submission, \cite{natekar2020representation}.
While \cite{natekar2020representation} present several proposed complexity measures, they explore accuracy on Mixed up and Manifold Mixed up training sets as potential predictors of generalization gap, and performance on mixed up data is one of the inputs into their winning submission.
\cite{natekar2020representation} mixup data points or intermediate representations within a class, not between classes.
While this closely resembles our work, in that the authors are using performance on a perturbed dataset, namely a mixed up one, the key difference is that \cite{natekar2020representation} only investigate a network's response to a single magnitude of interpolation, 0.5.
Additionally, we investigate \textit{between}-class interpolation as well.
Our proposed Gi-score and Pal-score therefore provide a much more robust sense for how invariant a network is to this mixing up perturbation and can easily be applied to other perturbations as well.

In the vein of exploring various transformations / perturbations, the second place submission \cite{kashyap2021robustness} perform various augmentations, such as color saturation, applying Sobel filters, cropping and resizing, and others, and create a composite penalty score based on how a network performs on these perturbed data points.
Our work, in addition to achieving better generalizatiton gap prediction scores, can be thought of as an extension of this submission, because as above, rather than looking at a single magnitude of various perturbations, the Gi-score and Pal-score provide a summary of how a model reacts to a spectrum of parameterized transformations.

\section{Methodology}\label{sec:methodology}
\subsection{Notation}\label{subsec:notation}

We begin by defining a network for a classification task as $f: \R^d \rightarrow \Delta_k$; that is, a mapping of real input signals $x$ of dimension $d$ to discrete distributions, with $\Delta_k$ being the space of all $k$-simplices.
We also define the intermediate layer mappings of a network as $f^{(\ell)}: \R^{d_{\ell-1}} \rightarrow \R^{d_{\ell}}$, where $\ell$ refers to a layer's depth with dimension $d_{\ell}$.
The output of each layer is defined as $x^{(\ell)} = f^{(\ell)}(x^{(\ell-1)})$, with inputs defined as $x^{(0)}$.
Additionally, let $f_{\ell}: \R^{d_\ell} \rightarrow \Delta_k$ be the function that maps intermediate representations $x^{(\ell)}$ to the final output of probability distributions over classes. 
For a dataset $\mathcal{D}$, consisting of pairs of inputs $x \in \R^d$ and labels $y \in [k]$, a network's accuracy is defined as $\mathcal{A} = \sum_{x, y \in \mathcal{D}}\one(\max_{i \in [k]}f(x)[i] = y) \ / \ |\mathcal{D}|,$
i.e. the fraction of samples where the predicted class matches the ground truth label, where $\one(\cdot)$ is an indicator function and $f(x)[i]$ refers to the probability weight of the $i^{\mathrm{th}}$ class.

We define perturbations of the network's representations as $\mathcal{T}_{\alpha}: \R^{d_{\ell}} \rightarrow \R^{d_{\ell}}$, where $\alpha$ controls the magnitude of the perturbation.
For example, adding Gaussian noise with zero mean and standard deviation $\sigma_\alpha$ to inputs can be represented as $\mathcal{T}_\alpha(x^{(0)}) = x^{(0)} + \epsilon$, where $\epsilon \sim \mathcal{N}(\mathbf{0}, \sigma_\alpha\mI).$
To measure a network's response to a perturbation $\mathcal{T}_\alpha$ applied at the $\ell^{\mathrm{th}}$ layer output, we calculate the accuracy of the network given the perturbation:
$\mathcal{A}_\alpha^{(\ell)} = \sum_{x, y \sim \mathcal{D}}\one(\max_{i \in [k]}f_{\ell}(\mathcal{T}_\alpha(x^{(\ell)}))[i] = y) \ / \ |\mathcal{D}|.$
The greater the gap $\mathcal{A} - \mathcal{A}_\alpha^{(\ell)}$, the less the network is resilient or invariant to the perturbation $\mathcal{T}_\alpha$ when applied to the $\ell^{\mathrm{th}}$ layer.  Perturbations at deeper network layers can be viewed as perturbations in an implicit feature space learned by the network.

\subsection{Calculating the Gi-score and Pal-score}\label{subsec:gi_score}
In our work, we let $\mathcal{T}_\alpha$ be defined as an interpolation between two points of either different or the same class:
$\mathcal{T}_\alpha(x) = (1-\alpha)x + \alpha x'$,
For \textit{inter}-class interpolation, i.e. where 
where $x'$ is a (random) input from a different class than $x$, we range $\alpha \in [0, 0.5)$.
For the \textit{intra}-class setup, i.e. where $x$ and $x'$ are drawn from the same class, we include the upper bound of the magnitude: $\alpha \in [0, 0.5].$
While we explored other varieties of perturbation, such as adding Gaussian noise, we found that this mixup perturbation was most predictive of generalization gaps for the networks and datasets we tested.
Both mixup perturbations that we tested (intra- and inter-class) are justifiable for predicting generalization gap.
We hypothesize that invariance to interpolation \textit{within} a class should indicate that a network produces similar representations and ultimately the same class maximum prediction for inputs and latent representations that are within the same class regions (captured by intra-class interpolation).
Invariance to interpolation \textit{between} classes up to 50\% should indicate that the network has well separated clusters for representations of different classes and is robust to perturbations moving points away from heavily represented class regions in the data / representations.

\begin{figure}[h]
    \centering
    \begin{subfigure}[b]{0.45\linewidth}
        \includegraphics[width=\linewidth]{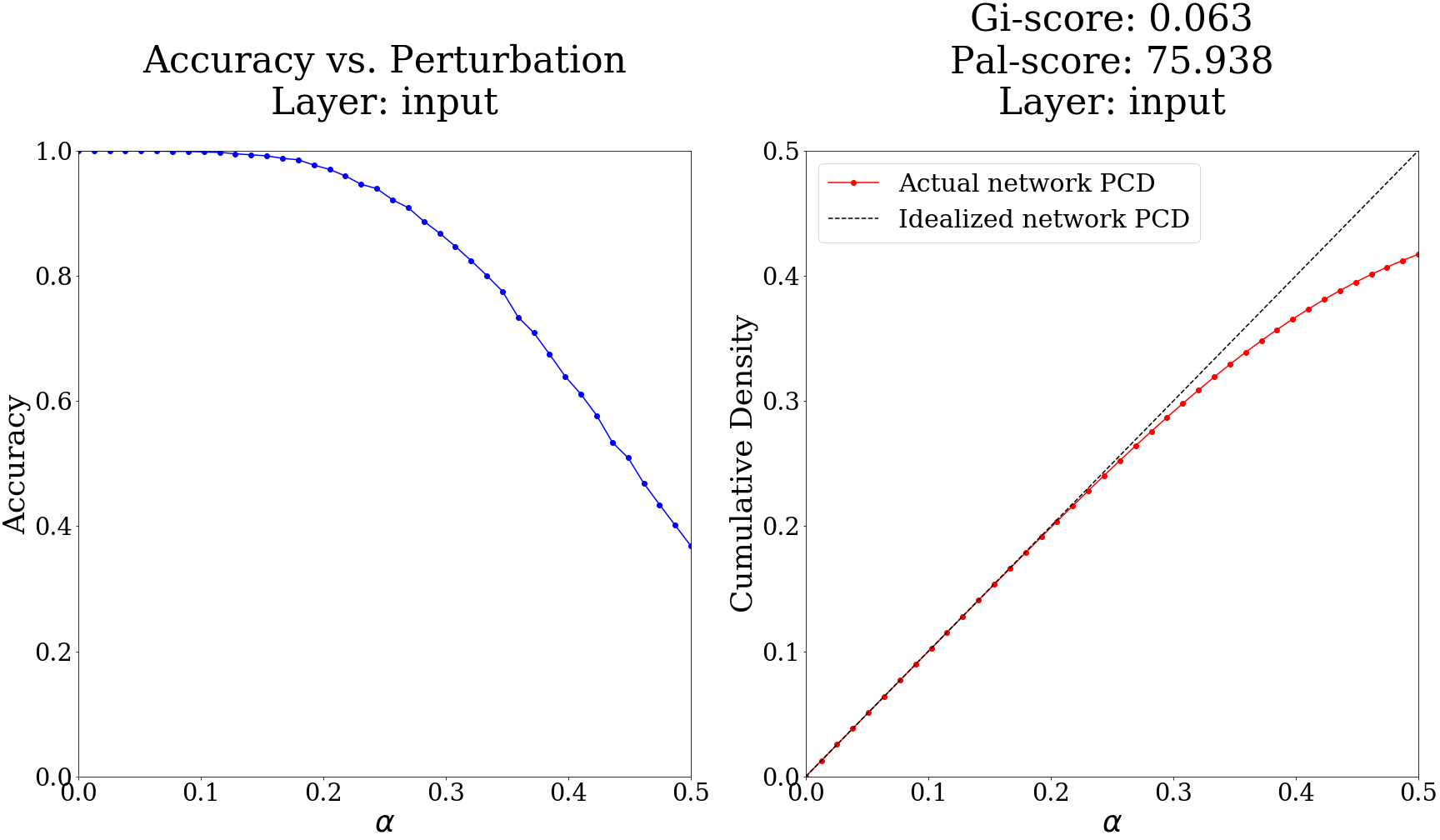}
    \end{subfigure}
    \begin{subfigure}[b]{0.45\linewidth}
        \includegraphics[width=\textwidth]{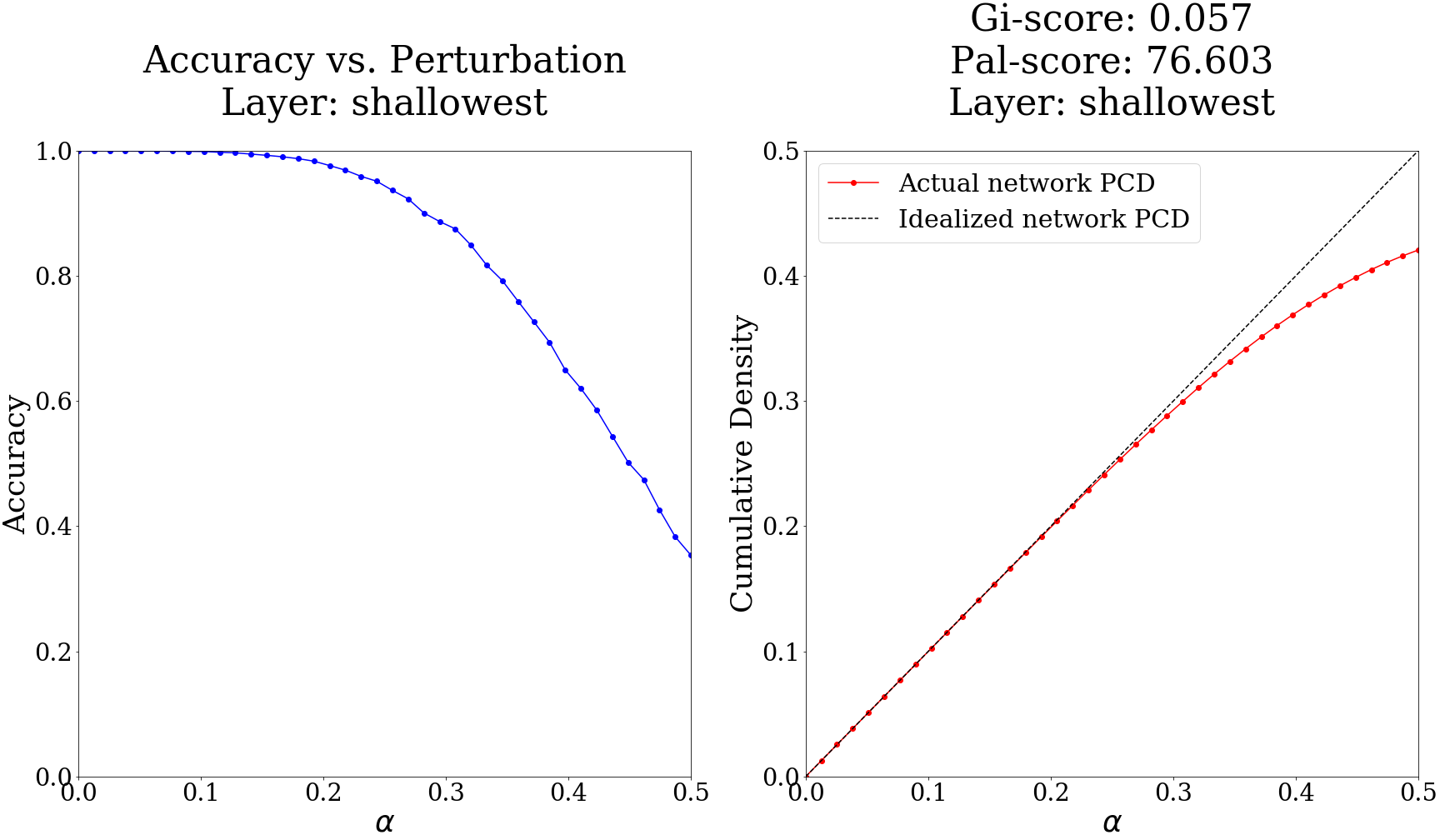}
    \end{subfigure}
    \caption{Sample Gi-score and Pal-score calculation for inter-class mixup applied to a VGG-like network trained on CIFAR-10 data.
    \textit{Perturbation-response curves} (blue) examine the network's response (accuracy on sample of training set) to varying levels of mixup perturbation on inputs: $x^{(0)}$ (left) and shallowest representations: $x^{(1)}$ (right).
    \textit{Perturbation-cumulative-density curves} (red) compare the network to idealized one that would produce 1.0 accuracy regardless of the magnitude of interpolation.}
    \label{fig:gi_scores}
\end{figure}

To measure a network's robustness to a perturbation, one could simply choose a fixed $\alpha$ and measure the network's response.
However, a more complete picture is provided by sampling the network's response to various magnitudes of $\alpha$.
In the blue plots in Figure \ref{fig:gi_scores}, we show this in practice.
For inter-class mixup ranging from 0 to 0.5, we measure the network's accuracy on a subset of the training set.
Only the inputs / layer outputs are mixed up, the label is held constant to test invariance to mixup.
We apply this perturbation at various depths, and in Figure \ref{fig:gi_scores}, we display perturbations at the input level ($x^{(0)}$) and shallowest layer's representation ($x^{(1)}$, the layer right after the input).
The results is the blue curves seen in Figure \ref{fig:gi_scores} that we refer to as \textit{perturbation-response} (PR) curves.

To extract a single statistic from the PR curves in Figure \ref{fig:gi_scores}, we draw inspiration from the Gini coefficient and Palma ratio, two distinct measures of income inequality that compare wealth distribution of a given economy with that of an idealized economy \citep{cobham2013all}.
Namely, we compare a network's response to varying magnitudes of perturbations with an \textit{idealized} network: one whose accuracy is unaffected by the perturbations.
The idealized network therefore has a PR curve that starts and remains at accuracy 1.0 regardless of the magnitude $\alpha < 0.5.$

This comparison is achieved by creating a new graph that plots the cumulative density integral under the PR curves against the magnitudes $\alpha_i \in [0, 0.5)$:
$\int_{0}^{\alpha_i} \mathcal{A}_\alpha d\alpha$.
This produces what we call \textit{perturbation-cumulative-density} (PCD) curves seen in red in Figure \ref{fig:gi_scores}.
For the idealized network whose PR is identically equal to 1 for all $\alpha$, this PCD curve is just the $45^{\circ}$ line passing through the origin.
Finally, the \textbf{Gi-score} (named for the Gini coefficient it draws inspiration from) is calculated by taking the area between the idealized network's PCD curve and that of the actual network.
The \textbf{Pal-score} (named for the Palma ratio it draws inspiration from) is calculated by dividing the area for the largest top 60\% of  perturbation magnitudes by the area for the bottom 10\%.
This allows us to focus on variations on the upper and lower ends of the perturbation magnitude spectrum, ignoring the middle perturbations that might not vary as widely across networks.


\section{Results}\label{sec:results}
We calculate our Gi-score and Pal-score on a corpus of trained networks and their corresponding datasets provided in \cite{jiang2020neurips}.
The networks from this competition span several architectures and datasets.
Namely, there are VGG, Network-in-Network (NiN), and Fully Convolutional (Full Conv) architectures.
The datasets are comprised of CIFAR-10 \citep{krizhevsky2009learning}, SVHN \citep{netzer2011reading}, CINIC-10 \citep{darlow2018cinic}, Oxford Flowers \citep{nilsback2008automated}, Oxford Pets \citep{parkhi2012cats}, and Fashion MNIST \citep{xiao2017fashion}.
The list of dataset-model combinations, or tasks, available in the trained model corpus can be seen in the first two rows of Table \ref{tab:results}.
Across the 8 tasks, there are a total of 550 networks.
Each network is trained so that it attains nearly perfect accuracy on the training dataset.
For each network, we also know the generalization gap.



As proposed in \cite{jiang2020neurips}, the goal is to find a \textit{complexity} measure of networks that is causally informative (predictive) of generalization gaps.
To measure this predictive quality, \cite{jiang2020neurips} propose a Conditional Mutual Information (CMI) score.
For full implementation details of this score, please reference \cite{jiang2020neurips}, but roughly, higher values of CMI represent greater capability of a complexity score in predicting generalization gaps.
In Table \ref{tab:results}, we present the average CMI scores for all models within a task for our Gi- and Pal-scores compared to that of the winning team \citep{natekar2020representation} from the PGDL competition.
We also compare our statistic to comparable ones presented in \cite{natekar2020representation} that rely on Mixup and Manifold Mixup accuracy\footnote{Scores come from \cite{natekar2020representation}.
For the scores not reported there, we use the code provided by the authors: https://github.com/parthnatekar/pgdl}.
The winning submission described in \cite{natekar2020representation} uses a combination of a score based on the accuracy of mixed up input data and a clustering quality index of class representations, known as the Davies-Bouldin Index (DBI) \citep{davies1979cluster}.
Using the notation introduced in Section \ref{sec:methodology}, the measures from \cite{natekar2020representation} present in Table \ref{tab:results} can be described as follows: Mixup accuracy: $\mathcal{A}_{0.5}^{(0)}$; Manifold Mixup accuracy: $\mathcal{A}_{0.5}^{(1)}$; DBI * Mixup: $DBI * (1-\mathcal{A}_{0.5}^{(0)}).$

\begin{table}[ht!]
\caption{Comparison of Conditional Mutual Information scores for various complexity measures across tasks.
The highest score within each task among measures that are only based on \textit{mixup} are bolded.
For reference, the PGDL winning measure of DBI*Mixup is included in the bottom row.
If the best mixup-based score out-performs the winning measure it is also marked with an asterisk.
For Gi- an Pal-scores, we indicate whether \textit{Inter} or \textit{Intra} mixup was used for the parametric perturbation along with the depth at which the perturbation was applied in parentheses, with input = 0 and shallowest layer = 1.
In the CINIC-10 columns, `bn' stands batch-norm.
}
\label{tab:results}
\begin{center}
\begin{tabular}{l | c c | c | c c | c | c | c}
\multicolumn{1}{c|}{}
&\multicolumn{2}{c|}{CIFAR-10}
&\multicolumn{1}{c|}{SVHN}
&\multicolumn{2}{c|}{CINIC-10}
&\multicolumn{1}{c|}{\makecell{Oxford \\ Flowers}}
&\multicolumn{1}{c|}{\makecell{Oxford \\Pets}}
&\multicolumn{1}{c}{\makecell{Fashion \\ MNIST}}
\\\cline{1-9}
\multicolumn{1}{c|}{}
&\textit{VGG}
&\textit{NiN}
&\textit{NiN}
&\makecell{\textit{Conv}\\\textit{w/bn}}
&\makecell{\textit{Conv}\\\textit{w/o bn}}
&\textit{NiN}
&\textit{NiN}
&\textit{VGG}
\\ \hline
Gi \textit{Inter} ($\ell=0$)  & 3.12        & \bf34.78$^*$ & 26.86        & 20.92        &  6.68        & 33.35        & \bf17.80$^*$ &  4.49\\
Gi \textit{Inter} ($\ell=1$)  & \bf7.69$^*$ & 24.02        & 12.25        & 12.62        &  8.42        &  7.39        &  4.57        & \bf16.12$^*$\\
Pal \textit{Inter} ($\ell=0$) & 3.17        & 27.79        & 22.91        & 20.94        &  6.21        & 29.75        & 15.96        &  4.16\\
Pal \textit{Inter} ($\ell=1$) & 7.10        & 13.33        &  9.65        & 12.11        &  7.69        &  6.27        &  3.49        & 14.43\\
\hline
Gi \textit{Intra} ($\ell=0$)  & 0.82        & 31.73        & \bf40.99$^*$ & 22.80        & 11.49        & \bf40.56     & 16.80        &  5.22\\
Gi \textit{Intra} ($\ell=1$)  & 0.23        & 16.82        & 10.98        &  9.40        & 12.38        &  6.85        &  3.49        &  5.74\\
Pal \textit{Intra} ($\ell=0$) & 0.66        & 24.64        & 29.77        & 24.38        & 10.93        & 38.04        & 15.25        &  4.93\\
Pal \textit{Intra} ($\ell=1$) & 0.45        & 10.25        & 14.08        &  8.80        & 10.65        &  5.96        &  3.02        &  6.25\\
\hline
Mixup                         & 0.03        & 14.18        & 22.75        & \bf30.30     & \bf19.51$^*$ & 35.30        & 9.99         & 7.75\\
Manifold Mixup                & 2.24        & 2.88         & 12.11        & 4.23         & 4.84         & 0.03         & 0.13         & 0.19\\
\hline \hline
\textit{DBI*Mixup$^1$}        & \it0.00     & \it25.86     & \it32.05     & \it31.79     & \it15.92     & \it43.99     & \it12.59     & \it9.24
\end{tabular}
\end{center}
\end{table}

These results highlight that the Gi-score and Pal-score perform competitively in predicting generalization gap.
Note that some versions of our scores out-perform the mixup approaches used in the PGDL winning approach on the majority of tasks, and even significantly out-perform the DBI*Mixup approach on 4 tasks.
This suggests the possibility of even better results combining our scores with DBI (future work).
In addition, we believe that our scores provide a more robust measure for how well a model is able to learn invariances to certain transformations.
For example, the Mixup complexity score presented in \cite{natekar2020representation} simply takes a 0.5 interpolation of data points and calculates accuracy of a network on the this mixed up portion of the training set.
In contrast, our scores allow us to capture network performance on a spectrum of interpolations, thereby providing a more robust statistic for how invariant a network is to linear data interpolation.
Our approach can also be extended to apply to any parametric transformation.

\section{Conclusion}\label{sec:conclusion}
In this work, we introduced two novel statistics inspired by income inequality metrics, which effectively predict neural network generalization.
The Gi-score and Pal-score are both computationally efficient, requiring only several forward passes through a subset of the training data, and intuitive.
Calculating these statistics on the corpus of data made available in \cite{jiang2020neurips} showed that the our scores applied to linear interpolation between data points have strong performance in predicting a network's generalization gap.

In addition to this predictive capability of generalization, we believe that the Gi-score and Pal-score provide a useful criterion for evaluating networks and selecting which architecture or hyperparameter configuration is most invariant to a desired transformation.
Because they rely on comparison with an idealized network that is unaffected by all magnitudes of a perturbation, future work will explore how these scores aid in discerning the extent to which a network has learned to be invariant to a given parametric transformation.



\subsubsection*{Acknowledgments}
The authors would like to thank the organizers of the Predicting Generalization in Deep Learning competition and workshop hosted at NeurIPS 2020 for providing a repository of pre-trained networks and their corresponding datasets and starter code for working with this corpus.

\bibliography{iclr2021_conference}
\bibliographystyle{iclr2021_conference}


\end{document}